\documentclass[conference]{IEEEtran}
\IEEEoverridecommandlockouts
\usepackage{cite}
\usepackage{amsmath,amssymb,amsfonts}
\usepackage{algorithmic}
\usepackage{graphicx}
\usepackage{textcomp}
\usepackage{xcolor}
\usepackage[hidelinks]{hyperref}

\def \methodname {AF Adapter}
\def \modelname {AF Adapter based RoBERTa}

\begin{document}

\title{\methodname{}: Continual Pretraining for Building Chinese Biomedical Language Model}

\author{
    \IEEEauthorblockN{Yongyu Yan}
    \IEEEauthorblockA{
        \textit{School of Info. Sci. \& Engi.} \\
        \textit{East China University of Sci. \& Tech.} \\
        Shanghai, China \\
        y30221069@mail.ecust.edu.cn
    }
    \and
    \IEEEauthorblockN{Kui Xue}
    \IEEEauthorblockA{
        \textit{\ \ \ \ \ \ \ \ \ \ \ \ \ \ Shanghai AI Lab.\ \ \ \ \ \ \ \ \ \ \ \ \ \ } \\
        Shanghai, China \\
        xuekui@pjlab.org.cn
    }
    \and
    \IEEEauthorblockN{Xiaoming Shi}
    \IEEEauthorblockA{
        \textit{\ \ \ \ \ \ \ \ \ \ \ \ \ \  Shanghai AI Lab.\ \ \ \ \ \ \ \ \ \ \ \ \ \ } \\
        Shanghai, China \\
        shixiaoming@pjlab.org.cn
    }
    \and
    \IEEEauthorblockN{Qi Ye}
    \IEEEauthorblockA{
        \textit{School of Info. Sci. \& Engi.} \\
        \textit{East China University of Sci. \& Tech.} \\
        Shanghai, China \\
        yeh\_qi1125@ecust.edu.cn
    }
    \and
    \IEEEauthorblockN{Jingping Liu}
    \IEEEauthorblockA{
        \textit{School of Info. Sci. \& Engi.} \\
        \textit{East China University of Sci. \& Tech.} \\
        Shanghai, China \\
        jingpingliu@ecust.edu.cn
    }
    \and
    \IEEEauthorblockN{Tong Ruan}
    \IEEEauthorblockA{
        \textit{School of Info. Sci. \& Engi.} \\
        \textit{East China University of Sci. \& Tech.} \\
        Shanghai, China \\
        ruantong@ecust.edu.cn
    }
}

\maketitle

\begin{abstract}
  Continual pretraining is a popular way of building a domain-specific pretrained language model from a general-domain language model.
  In spite of its high efficiency, continual pretraining suffers from catastrophic forgetting, which may harm the model's performance in downstream tasks.
  To alleviate the issue, in this paper, we propose a continual pretraining method for the BERT-based model, named Attention-FFN Adapter.
  Its main idea is to introduce a small number of attention heads and hidden units inside each self-attention layer and feed-forward network.
  Furthermore, we train a domain-specific language model named \modelname{} for the Chinese biomedical domain.
  In experiments, models are applied to downstream tasks for evaluation.
  The results demonstrate that with only about 17\% of model parameters trained, \methodname{} achieves 0.6\%, 2\% gain in performance on average, compared to strong baselines.
  Further experimental results show that our method alleviates the catastrophic forgetting problem by 11\% compared to the fine-tuning method.
  Code is available at \url{https://github.com/yanyongyu/AF-Adapter}.
\end{abstract}

\begin{IEEEkeywords}
Continual pretraining, Chinese biomedical natural language processing, Adapter tuning
\end{IEEEkeywords}

\section{Introduction}

Currently, a vast volume of Chinese biomedical literature emerges, including medical diagnosis dialogues on Chinese medical communities and medical knowledge in Chinese encyclopedias.
As an example, DXY\footnote{https://portal.dxy.cn/} (a Chinese medical community) contains millions of medical dialogues between 5.5 million patient users and 2 million doctor users, while Baidu Encyclopedia\footnote{https://baike.baidu.com/} (a Chinese encyclopedia) includes over 30 million articles, covering a wide range of topics related to health, disease, and medical treatments.
As a result, there is a growing demand for precise Chinese biomedical text mining tools to effectively extract valuable information from Chinese medical literature.

\textbf{N}atural \textbf{l}anguage \textbf{p}rocessing (NLP) methods greatly improve the automatic text mining from Chinese literature accurately.
These methods have undergone several stages, including algorithms based on small-scale expert knowledge~\cite{manning1999foundations}, shallow machine learning algorithms~\cite{liddy2001natural}, and deep learning algorithms~\cite{bib:bioner}\cite{bib:recursive_chemprot}.
Among these deep learning algorithms, recently, \textbf{p}retrained \textbf{l}anguage \textbf{m}odels (PLMs)~\cite{bib:bert}\cite{bib:roberta} have become increasingly popular, as they demonstrate impressive performance.
Specifically, PLMs are trained on a vast amount of text data to learn better representations of natural language, providing a strong foundation for downstream NLP tasks.

However, directly applying PLMs on general domains to biomedical text mining suffers from the word distribution bias between general and biomedical texts, which harms performances.
To alleviate the issue, PLMs in the biomedical domain are introduced.
Methods of PLMs on specific domains are divided into two ways, pretraining from scratch~\cite{bib:pubmedbert} and continual pretraining~\cite{bib:biobert} based on general-domain language models.
Pretraining from scratch means training directly on a specialized corpus with a specialized vocabulary.
Simultaneously, continual pretraining first initialize with PLMs on general-domain and then resume pre-training on a domain-specialized corpus.
Compared with pretraining from scratch, continual pretraining benefits from higher training efficiency.
In this work, a continual pretraining manner is adopted.

\begin{figure}[!t]
  \centering
  \includegraphics[width=0.9\columnwidth]{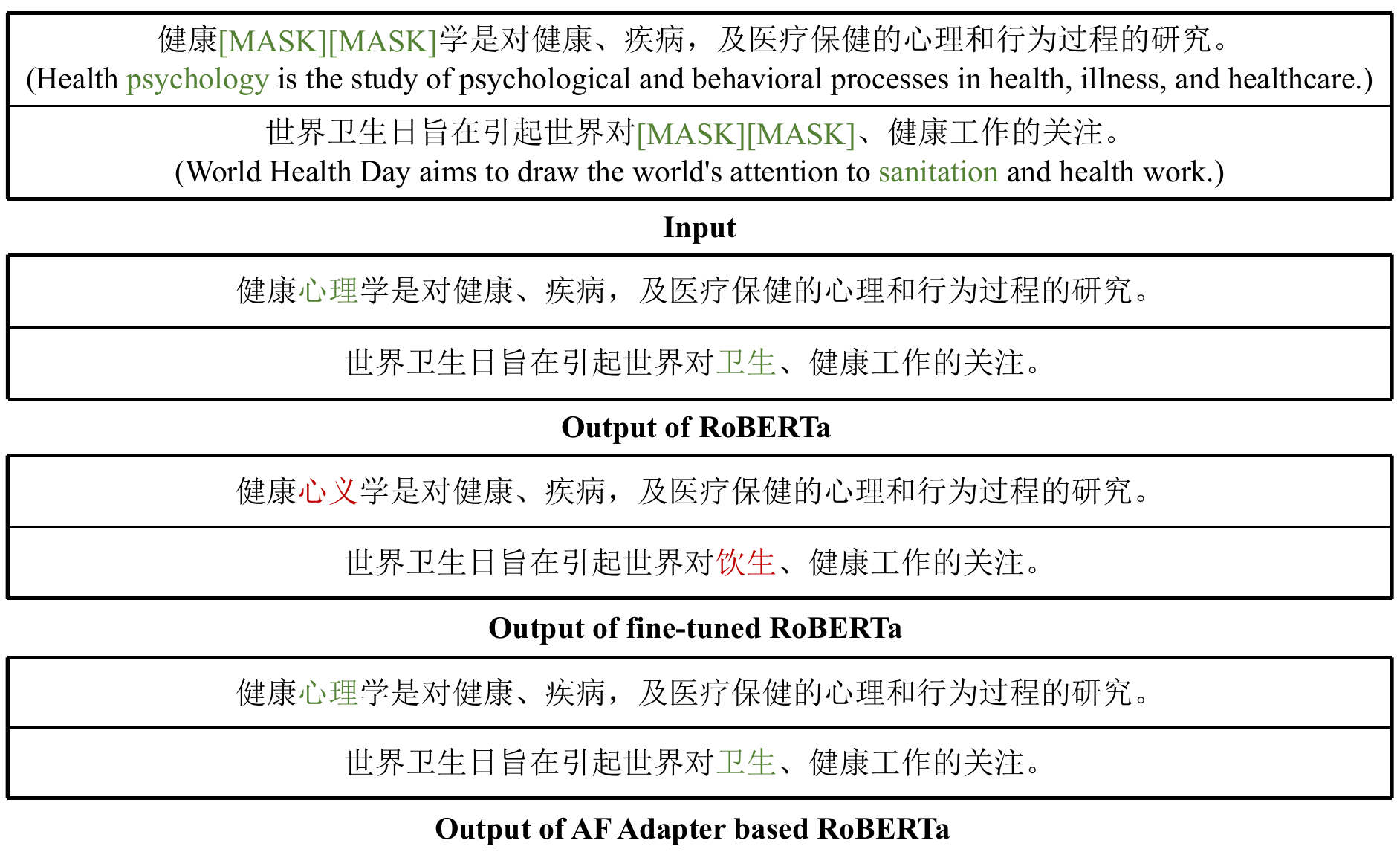}
  \caption{Outputs of RoBERTa, fine-tuned RoBERTa, and \modelname{} for two input items.}
  \label{fig:forgetting}
\end{figure}

In spite of the high training efficiency, continual pretraining suffers from catastrophic forgetting~\cite{bib:catastrophic-forgetting}\cite{bib:catastrophic-forgetting-2}\cite{bib:catastrophic-forgetting-3}.
Catastrophic forgetting means the model trained on new corpora tends to forget the knowledge of previous data.
For example, in Fig.~\ref{fig:forgetting}, fine-tuned RoBERTa encounters catastrophic forgetting, making mistakes to predict masked words.
Catastrophic forgetting of commonsense knowledge harms performance.

To alleviate the issue, there are three main approaches.
The first one is training-based methods, including discriminative fine-tuning, slanted triangular learning rates, and gradual unfreezing~\cite{bib:ulmfit}.
The second one is parameter reserve~\cite{bib:mixout}, which keeps randomly selected pretrained parameters.
The third one is adapter-based tuning methods~\cite{bib:adapter}\cite{bib:adapter-forgetting}, which inserts additional layers after specific layers of pretrained models.
Adapter-based tuning methods better mitigate the forgetting problem than other methods, thus becoming popular recently.

Despite the good performance, the adding-layer manner of the current adapter-based tuning methods adds depth to the networks.
The input needs to be fed forward to more layers and thus is more likely to be forgotten.
To alleviate the issue, we propose a layer-extending continual pretraining method, named \textbf{A}ttention-\textbf{F}FN \textbf{Adapter} (\methodname{}), which extends the attention matrixes and \textbf{f}eed-\textbf{f}orward \textbf{n}etworks (FFN).
Specifically, its essential idea is to introduce a small number of heads and hidden units inside each self-attention and FFN layer of BERT.
In this method, only the added parameters are trainable, and the original pretrained parameters are frozen.
This layer-extension method does not increase network layer depth, thus alleviating the input information forgetting problem.
Then, \modelname{} is obtained by fine-tuning a pretrained language model on the medical domain with \methodname{}.
To estimate the feasibility of \methodname{}, \modelname{} is compared with other models in the downstream NLP tasks.
Experimental results demonstrate that with only 17\% of model parameters trained, \modelname{} achieves 0.6\%, 2\% gain in performance compared with state-of-the-art models, on average.

\textbf{Contributions.}
The contributions in this paper are summarized as follows:

\begin{itemize}
  \item We propose a layer-extending continual pretraining method called \methodname{} (Attention-FFN Adapter), which aims to extend attention heads and hidden units for the BERT-based model.
        The method further alleviates catastrophic forgetting compared with layer-adding methods.
  \item We propose a Chinese biomedical-domain pretrained language model called \modelname{}, which is trained using \methodname{} and biomedical corpus.
        \modelname{} contributes greatly to downstream medical tasks.
  \item Our method achieves 0.6\%, 2\% gain in performance on average, compared to strong baselines on the CBLUE benchmark.
        Besides, \methodname{} alleviates the catastrophic forgetting problem by 11\% compared to the fine-tuning method.
\end{itemize}

\section{Related Work}

\begin{figure*}[!t]
  \small
  \centering
  \includegraphics[width=0.9\textwidth]{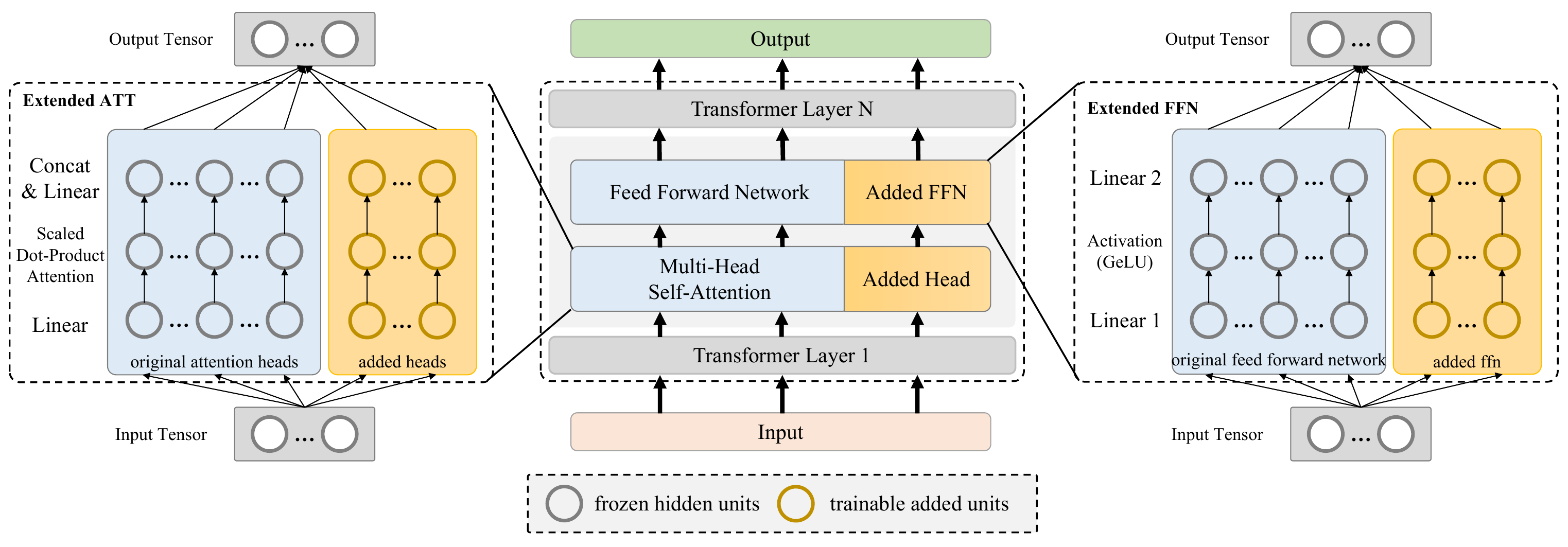}
  \caption{
    The illustration of \methodname{}.
    \methodname{} introduces a small number of additional heads and hidden units to each self-attention layer and feed-forward network of the BERT-based model.
    The added parameters of each layer are independent.
    The heads and hidden units with orange color are trainable, and all parameters with grey color from the original model are frozen.}
  \label{fig:overview}
\end{figure*}

In this section, we provide a brief overview of the existing approaches in the fields of language modeling and domain-specific pretraining.
We present a summary of different methods and techniques that have been employed in these areas.

\subsection{Language Modeling}

In natural language processing, pretraining language models with large amounts of unannotated data have proven to be a successful strategy for transfer learning.
It is effective to learn general language representations from language models and transfer the model to specific downstream tasks by fine-tuning.

Language models are usually trained on particular pretraining tasks using encyclopedic and newsletter corpus.
The pretraining tasks are crucial for learning general language representations.
For self-supervised pretraining, the widely-used tasks are as follows:

\textbf{Language Modeling (LM)} is a classic probabilistic density estimation problem, used in e.g., GPT-2~\cite{bib:gpt-2}, ULMFiT~\cite{bib:ulmfit} and SiATL~\cite{bib:siatl}.
The next output token depends on the joint probability of the previous tokens.
A drawback of unidirectional language modeling is that the representation of each token encodes only the leftward context tokens.
However, better contextual representations of text should encode contextual information from both directions.

\textbf{Masked Language Modeling (MLM)} is a pretraining task to overcome the drawback of unidirectional language modeling.
MLM first masks out some tokens in the input text and then trains the model to predict the masked tokens.
MLM is used in BERT~\cite{bib:bert}, MASS~\cite{bib:mass}, T5~\cite{bib:t5}, RoBERTa~\cite{bib:roberta}, etc.

\textbf{Permuted Language Modeling (PLM)} is a language modeling task on a random permutation of the input text, used by XLNet~\cite{bib:xlnet}, BART~\cite{bib:bart}.
They observed that certain special tokens, such as [MASK], are not present during downstream tasks.
To bridge the gap between pre-training and fine-tuning, the model is trained to predict some tokens in permuted token sequence, depending on the rest of the tokens.

\textbf{Contrastive Learning (CTL)} assumes some observed pairs of text which are more semantically similar than randomly sampled text.
Common CTL tasks include Next Sentence Prediction (NSP)~\cite{bib:bert}\cite{bib:nsp}, Sentence Order Prediction (SOP)~\cite{bib:sop}\nolinebreak\cite{bib:albert}.

\subsection{Domain-Specific Pretraining}

Specialized domains like biomedicine pose challenges for general-domain pretrained language models due to the following reasons:

\begin{itemize}
  \item General-domain pretrained language models are trained on datasets primarily sourced from encyclopedias and newsletters, making it difficult to estimate their performance on biomedical domain text.
  \item Word distributions differ significantly between general-domain text and biomedical-domain text, including the presence of biomedical terms.
  \item Biomedical text in Chinese biomedicine exhibits complex phrase combinations and structures.
\end{itemize}

To address these challenges, researchers have proposed studies focusing on biomedical domain-specific pretraining, which can be categorized into two main types.

\textbf{Domain-specific pretraining from scratch} is based on large-scale biomedical corpora.
For instance, PubMedBERT~\cite{bib:pubmedbert} and BioGPT~\cite{bib:biogpt} are trained on PubMed-based corpora, containing 3.1 billion words and 15 million items respectively.
One notable advantage of domain-specific pretraining from scratch is the ability to customize the model's vocabulary to the domain.
This allows for more appropriate tokenization of medical terms, avoiding fragmented subword representations.

\textbf{Continual pretraining of a general-domain pretrained model} is a common approach for pretraining a biomedical model.
This approach involves initializing with a standard model pretrained on encyclopedia and newsletter corpora and then continuing the pretraining process using biomedical corpora.
Continual pretraining benefits from the knowledge acquired during general-domain pretraining.
Recent studies have demonstrated that injecting extra knowledge information can enhance the model~\cite{bib:knowledge-enhance}, such as knowledge acquisition~\cite{bib:knowledge-acquisition}.
However, sequential task training may lead to catastrophic forgetting, where the model forgets the general-domain knowledge~\cite{bib:catastrophic-forgetting}\cite{bib:catastrophic-forgetting-2}\cite{bib:catastrophic-forgetting-3}.
BioBERT is an example of this approach, where continual pretraining is initialized with weights from BERT and conducted using the PubMed corpus~\cite{bib:biobert}.
In the Chinese context, PCL-MedBERT\footnote{\url{https://www.ihub.org.cn/html/2020/news\_0824/47.html}} is trained based on BERT using biomedical text and medical QA corpora.

\section{Preliminaries}

In this section, we provide a concise review of two key components in the BERT architecture: the multi-head self-attention mechanism and the feed-forward network.
We briefly explain their roles and functions within the BERT model.

\subsection{BERT}

The BERT model architecture is built upon a multi-layer bidirectional Transformer encoder structure~\cite{bib:transformer}.
Each encoder consists of a stack of identical blocks that incorporate multi-head self-attention and feed-forward networks. The multi-head self-attention layer can be formulated as follows:

\begin{equation}
  \label{bert-qkv}
  \begin{aligned}
     & Q(\boldsymbol{x}) = \boldsymbol{x}W_Q + b_Q, \\
     & K(\boldsymbol{x}) = \boldsymbol{x}W_K + b_K, \\
     & V(\boldsymbol{x}) = \boldsymbol{x}W_V + b_V,
  \end{aligned}
\end{equation}
where $W_Q \in \mathbb{R}^{d_{model} \times hd_k}$, $W_K \in \mathbb{R}^{d_{model} \times hd_k}$, $W_V \in \mathbb{R}^{d_{model} \times hd_v}$, $b_Q \in \mathbb{R}^{d_k}$, $b_K \in \mathbb{R}^{d_k}$ and $b_V \in \mathbb{R}^{d_{v}}$ represent the weight matrices and biases.
$d_{k}=d_{v}=d_{model}/h$, $d_{model}$ is the model dimension, and $h$ is the number of attention heads.
The $Q$, $K$, and $V$ are then split into $h$ parts to calculate the multi-head attention.
This can be expressed as:

\begin{equation}
  \label{bert-attention}
  \begin{aligned}
     & head_j = \text{Attention}(Q_j, K_j, V_j),                      \\
     & \text{MultiHead}(\boldsymbol{x}) = [head_1, \dots, head_h]W_O, \\
  \end{aligned}
\end{equation}
where $Q = [Q_1:\dots:Q_h]$, $K = [K_1:\dots:K_h]$, $V = [V_1:\dots:V_h]$ and $W_O \in \mathbb{R}^{hd_v \times d_{model}}$.
The operation ``$:$'' stand for the column-wise concatenation.

After the multi-head attention, the feed-forward network is employed, consisting of two linear transformations with a GeLU activation function in between:

\begin{equation}
  \label{bert-ffn}
  \text{FFN}(\boldsymbol{x}) = \text{GeLU}(\boldsymbol{x}W_1 + b_1)W_2 + b_2,
\end{equation}
where $W_1 \in \mathbb{R}^{d_{model} \times d_{ff}}$, $W_2 \in \mathbb{R}^{d_{ff} \times d_{model}}$, $b_1 \in \mathbb{R}^{d_{ff}}$ and $b_2 \in \mathbb{R}^{d_{model}}$ represent the weight matrices and biases applied to the input $\boldsymbol{x}$ in the feed-forward network.

\section{Methodology}

In this section, \methodname{} is introduced to train domain-specific PLMs based on general-domain PLMs.
The overall architecture of \methodname{} is illustrated in Fig.~\ref{fig:overview}.

\subsection{Model Architecture}\label{method:architecture}

Following adapter tuning~\cite{bib:adapter}\cite{bib:fl-tuning}, to alleviate catastrophic forgetting in the fine-tuning stage, additional trainable domain-specific parameters are inserted into the extension of the attention and FFN layer, while parameters in the original model are fixed to preserve the knowledge of the general domain.

\begin{figure}[!t]
  \small
  \centering
  \includegraphics[width=0.95\columnwidth]{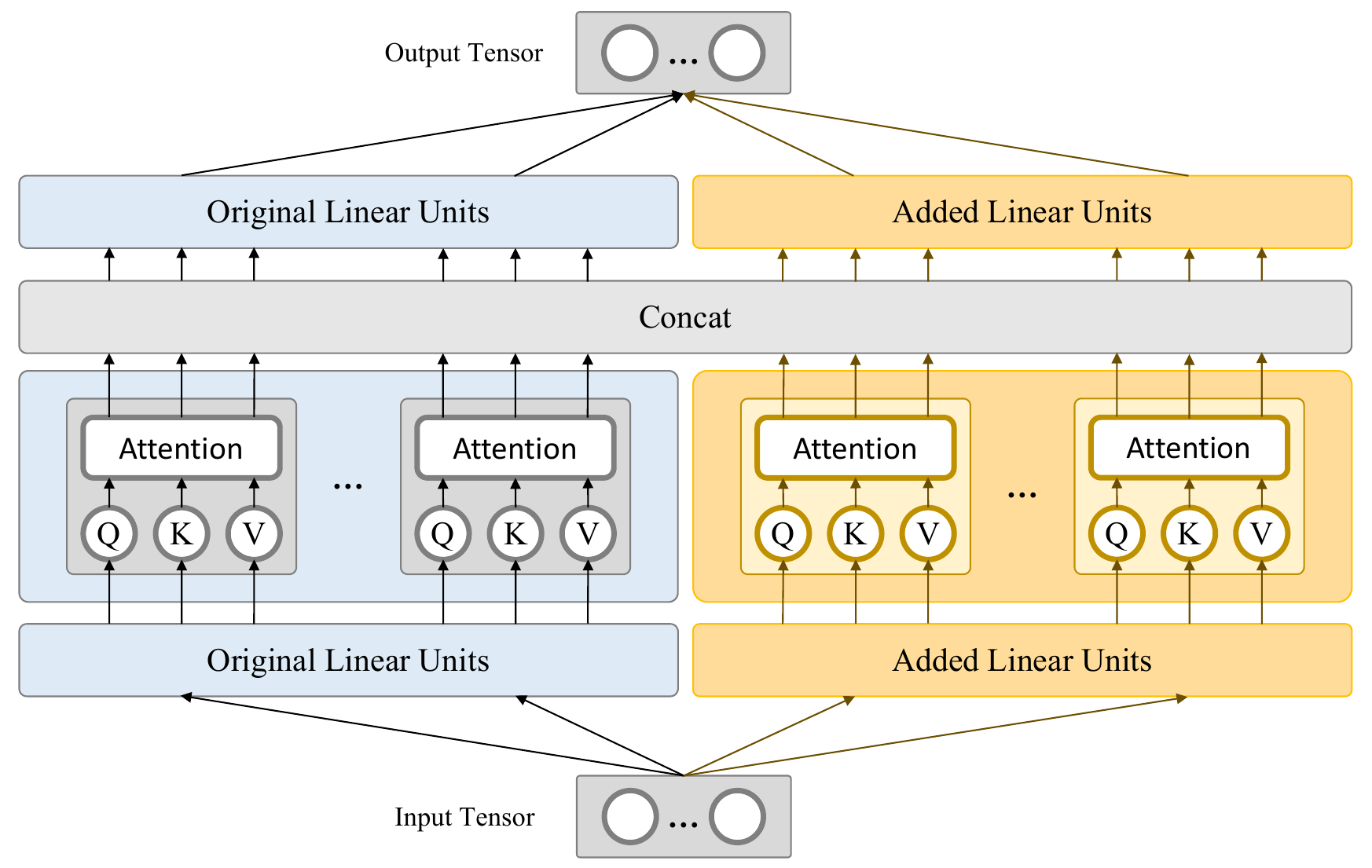}
  \caption{The attention layer architecture of \methodname{}.}
  \label{fig:attention}
\end{figure}

\textbf{Extended Attention Layer.}
A key component of BERT is the multi-head scaled dot-product self-attention mechanism.
In this work, the self-attention layer is extended to alleviate catastrophic forgetting.
Specifically, as shown in Fig.~\ref{fig:attention}, a number of ``domain-specific heads'' in each attention layer are added.
Note that added domain-specific heads in different layers are independent of each other.
To extend the self-attention mechanism, $i$ additional attention heads are added.
Formally, the input into the multi-head self-attention layer is denoted as $\boldsymbol{x}$.
$\boldsymbol{x}$ is firstly fed into linear layers, and mapped into $Q'$, $K'$, $V'$.
The extended weight matrices are denoted as $[W_Q:W'_Q]$, $[W_K:W'_K]$, and $[W_V:W'_V]$. $W_{Q}$, $W_{K}$, and $W_{V}$ are matrices from the original pretrained model, as described in Equation~\ref{bert-qkv}.
Then, $Q'(\boldsymbol{x}) \in \mathbb{R}^{d_{model} \times (h + i)d_{k}}$, $K'(\boldsymbol{x}) \in \mathbb{R}^{d_{model} \times (h + i)d_{k}}$, $V'(\boldsymbol{x}) \in \mathbb{R}^{d_{model} \times (h + i)d_{v}}$ are obtained,
\begin{equation}
  \begin{aligned}
    & Q'(\boldsymbol{x}) = \boldsymbol{x}[W_Q:W'_Q] + [b_Q:b'_Q],     \\
    & K'(\boldsymbol{x}) = \boldsymbol{x}[W_K:W'_K] + [b_K:b'_K],     \\
    & V'(\boldsymbol{x}) = \boldsymbol{x}[W_V:W'_V] + [b_V:b'_V],
  \end{aligned}
\end{equation}
where the projections are parameter matrices $W'_{Q} \in \mathbb{R}^{d_{model} \times id_{k}}$, $W'_{K} \in \mathbb{R}^{d_{model} \times id_{k}}$, $W'_{V} \in \mathbb{R}^{d_{model} \times id_{v}}$, $b'_Q \in \mathbb{R}^{id_{k}}$, $b'_K \in \mathbb{R}^{id_{k}}$ and $b'_V \in \mathbb{R}^{id_{v}}$. $d_{k}$, $d_{v}$, $d_{model}$, $h$ are hyperparameters described in Equation~\ref{bert-qkv}.

Then, the $Q'$, $K'$ and $V'$ are divided into $h + i$ parts for calculating the multi-head attention,
\begin{equation}
  \begin{aligned}
                     & head'_{j} = Attention(Q'_{j}, K'_{j}, V'_{j}), \\
    \textrm{where}\  & Q' = [Q_1:\dots:Q_h:Q'_1:\dots:Q'_i],          \\
                     & K' = [K_1:\dots:K_h:K'_1:\dots:K'_i],          \\
                     & V' = [V_1:\dots:V_h:V'_1:\dots:V'_i].
  \end{aligned}
\end{equation}

Finally, the outputs of these heads are concatenated and then fed to linear transformations,
\begin{equation}
  \begin{aligned}
     & \rm{MultiHead}(\boldsymbol{x})                                           \\
     & = [head_1, ..., head_h, head'_1, ..., head'_i][W_O \bot W'_O],
  \end{aligned}
\end{equation}
where $W'_O \in \mathbb{R}^{(d_{k} \times i) \times d_{model}}$, and $head_{j}$ is the $j$th head described in Equation~\ref{bert-attention}. The operation ``$\bot$'' represents the row-wise concatenation.

\begin{figure}[!t]
  \centering
  \includegraphics[width=0.95\columnwidth]{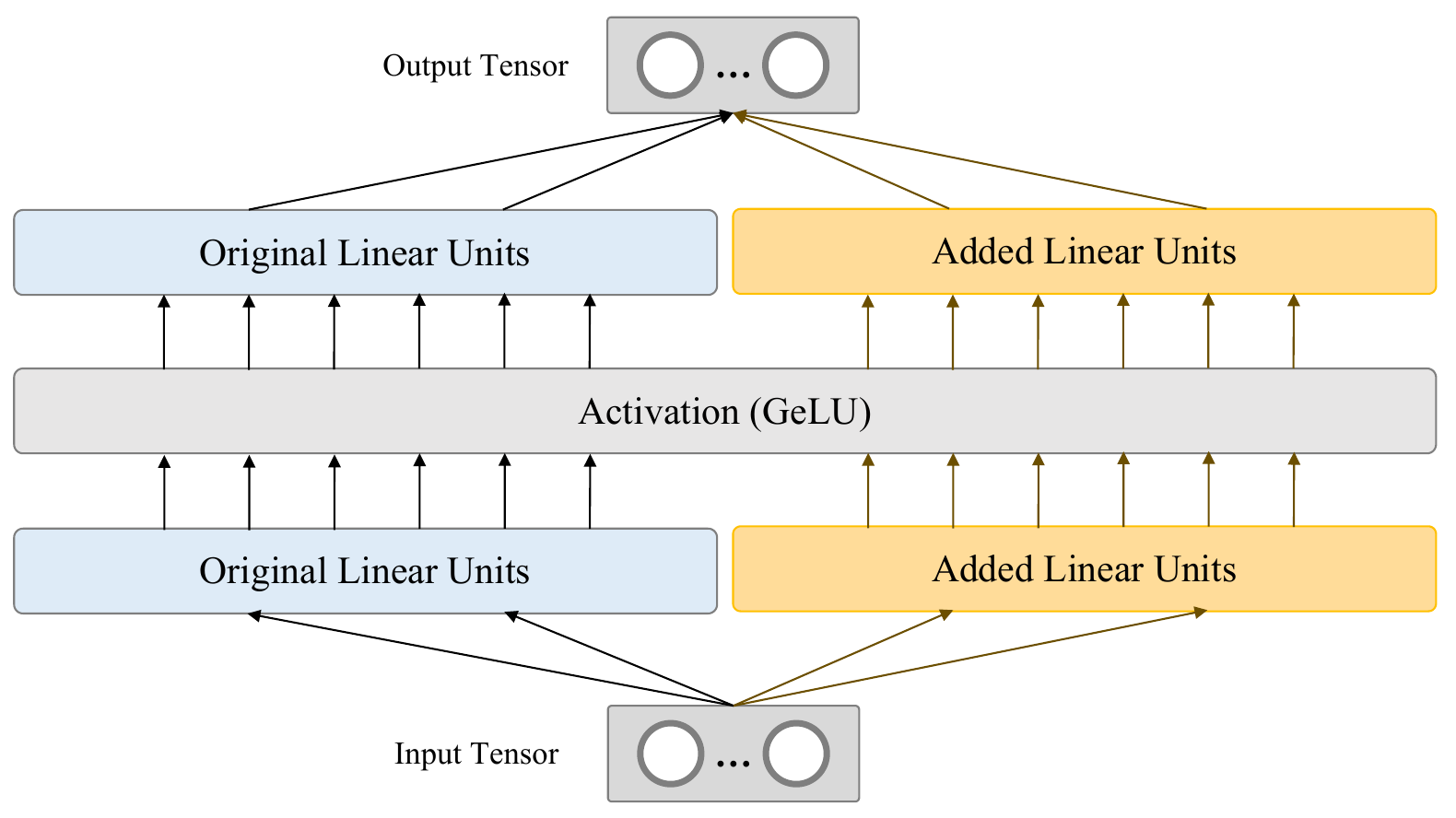}
  \caption{The feed-forward network architecture of \methodname{}.}
  \label{fig:ffn}
\end{figure}

\textbf{Extended Feed-Forward Network.}
For FFN in the model, a multi-layer perceptron with one hidden layer, we add a number of hidden units in each FFN layer with GeLU activation between two linear transformations, as shown in Fig.~\ref{fig:ffn}.
Formally, we add ``domain-specific hidden units'' of size $a$ by extending $W_1$ and $W_2$ to $[W_1:W'_1]$ and $[W_2 \bot W'_2]$,
\begin{equation}
  \begin{aligned}
     & \rm{FFN}(\boldsymbol{x})                                                                   \\
     & = \rm{GeLU}(\boldsymbol{x}[W_1:W'_1] + [b_1:b'_1])[W_2 \bot W'_2] + b_2 + b'_2,
  \end{aligned}
\end{equation}
where $W'_1 \in \mathbb{R}^{d_{model} \times a}$, $b'_1 \in \mathbb{R}^{a}$, $W'_2 \in \mathbb{R}^{a \times d_{model}}$, and $b'_2 \in \mathbb{R}^{d_{model}}$. $W_1$, $W_2$, $b_1$ and $b_2$ are metrices described in Equation~\ref{bert-ffn}.

\subsection{Pretraining Details}

\textbf{Pretraining Tasks.}
To continually pretraining the BERT-based model, the \textbf{m}asked \textbf{l}anguage \textbf{m}odeling (MLM) is utilized to pretrain the model.
In MLM, a subset of input tokens is randomly replaced with a special token (e.g., \texttt{[MASK]}), and MLM is designed to predict these tokens.
The training objective is the cross-entropy loss between the original tokens and the predicted ones.
As same as BERT and RoBERTa, 15\% of the input tokens are chosen, among which a random 80\% are replaced by \texttt{[MASK]}, 10\% are left unchanged, and 10\% are randomly replaced by a token from the vocabulary.

\textbf{Whole Word Masking.}
In the original BERT, the text is split into subwords and tokenized into tokens.
The whole word masking mitigates the drawback of masking only a part of the whole word.
For the whole word masking in Chinese, the traditional Chinese word segmentation tool ``Jieba'' is utilized to split the sentence into several words and provide extra word information to the data collator before masking.

\textbf{Training Strategy.}
Fine-tuning with domain-specific corpora is a standard method to continually pretrain the general-domain models.
However, this method suffers from the catastrophic forgetting problem~\cite{bib:catastrophic-forgetting}.

To alleviate the issue, we first make progress in the model architecture and do not train all parameters in the model.
Only the domain-specific parameters added in the self-attention and FFN ($W'_Q$, $W'_K$, $W'_V$, $W'_O$, $W'_1$, $W'_2$) are trained and all parameters inherited from the general-domain model are frozen.

Corpora are preprocessed into datasets before training.
Paragraphs are split into sentences by separators, and sentences are grouped to the max sequence length of the model.

\begin{table}[!t]
    \caption{Statistics of corpora for model pretraining.\label{table:datasets}}
    \renewcommand\arraystretch{1.2}
    \begin{center}
        \begin{tabular}{|c|c|c|}
            \hline
            \textbf{Corpus Type}          & \textbf{\# of Sentences}  & \textbf{\# of Tokens} \\
            \hline
            Biomedical Question Answering & 4,842k                    &  93M      \\
            Medical Encyclopedia          & 793k                      &  33M      \\
            Electronic Medical Record     & 1,762k                    &  45M      \\
            \hline
        \end{tabular}
    \end{center}
\end{table}

\section{Experiments}

In this section, we provide the details of our experimental setup and present the results obtained from the evaluation.
We pretrain the domain-specific models using collected datasets and then evaluate them on downstream tasks from the CBLUE~\cite{bib:cblue} benchmark.
Furthermore, we conduct an ablation study on different pretraining techniques and investigate convergence and stability.
Additionally, we perform a catastrophic forgetting analysis to evaluate the models' ability to retain previously learned knowledge when trained on new tasks.

\subsection{Pretraining Data and Settings}

\textbf{Datasets}.
To evaluate the performance of our method, we collect a variety of Chinese biomedical corpora.
These corpora include Chinese biomedical question answering, Chinese medical encyclopedia from Baidu Encyclopedia and Wikipedia, and Electronic Medical Records from Fudan University Shanghai Cancer Center.
Details of the pretraining corpora used for our experiments are presented in Table~\ref{table:datasets}.
We use these corpora to pretrain our model before fine-tuning on evaluation tasks.
By leveraging these diverse and comprehensive datasets, we aim to improve the model's ability to understand natural language text in the biomedical domain.

\textbf{Backbone.}
We perform experiments based on the Chinese RoBERTa-wwm-ext-base model~\cite{bib:chinese-roberta}, which is obtained from the HuggingFace Hub.
Specifically, the model originally consists of 12 transformer encoder layers, each with a hidden size of 768.
Each layer also includes 12 self-attention heads and a feed-forward network with an intermediate size of 3,072.
We add one additional attention head to each layer and increase the intermediate size of the feed-forward network by 1,024.
We transfer all the model parameters from the Chinese RoBERTa-wwm-ext-base model to our modified architecture.

\textbf{Pretraining Settings.}
We pretrain our model using the original vocabulary provided with the Chinese RoBERTa model.
We use the AdamW optimizer with warm-up and weight decay.
Specifically, we begin with a learning rate of zero, which increases linearly to a peak rate of $4 \times 10^{-4}$ over the first 1000 steps of training.
The learning rate then decays linearly to zero over the remaining steps.
We train the model for 100,000 steps using a total batch size of 512, spread across two NVIDIA A100 (80G) GPUs with a batch size of 64 and gradient accumulation steps of 4.
We use the Chinese whole-word masking (WWM) technique during pretraining, with a masking rate of 15\% and a maximum sequence length of 512 tokens.

\textbf{Baselines.}
The baseline models used in our experiment are of similar sizes and have been widely used.
For comparison of PLMs, the models include BERT-base~\cite{bib:bert}, BERT-wwm-ext-base~\cite{bib:chinese-roberta}, RoBERTa-wwm-ext-base~\cite{bib:chinese-roberta}, PCL-MedBERT, and MacBERT-base~\cite{bib:macbert}.
All these models have undergone pre-training on large-scale corpora.
For comparison of pretraining techniques, the baselines include Fine-Tuning, FL-Tuning~\cite{bib:fl-tuning}, and LoRA~\cite{bib:lora}.
We compare the performance of our model against these baselines to evaluate its effectiveness.

\subsection{Evaluation Tasks}

\begin{table}[t]
    \caption{Statistics of CBLUE tasks.\label{table:finetuning_tasks}}
    \renewcommand\arraystretch{1.2}
    \begin{center}
        \begin{tabular}{|c|c|c|c|c|c|}
            \hline
            \textbf{Task} & \textbf{Type} & \textbf{Train} & \textbf{Dev} & \textbf{Test} & \textbf{Metric} \\
            \hline
            CMeEE         & NER           & 15,000          & 5,000         & 3,000          & Micro-F1        \\
            CMeIE         & RE            & 14,339          & 3,585         & 4,482          & Micro-F1        \\
            CHIP-CDN      & NORM          & 6,000           & 2,000         & 10,000         & Micro-F1        \\
            CHIP-STS      & TS            & 16,000          & 4,000         & 10,000         & Macro-F1        \\
            CHIP-CTC      & TC            & 22,962          & 7,682         & 10,192         & Macro-F1        \\
            KUAKE-QIC     & TC            & 6,931           & 1,955         & 1,994          & Acc             \\
            KUAKE-QTR     & NLI           & 24,174          & 2,913         & 5,465          & Acc             \\
            KUAKE-QQR     & NLI           & 15,000          & 1,599         & 1,596          & Acc             \\
            \hline
        \end{tabular}
    \end{center}
\end{table}

\begin{table*}[t]
    \caption{Comparison results of different models on CBLUE tasks.\label{table:benchmark_results}}
    \renewcommand\arraystretch{1.2}
    \begin{center}
        \begin{tabular}{|l|c|c|c|c|c|c|c|c|c|}
            \hline
            \textbf{Model}       & \textbf{CMeEE}  & \textbf{CMeIE}  & \textbf{CDN}  & \textbf{CTC}  & \textbf{STS}  & \textbf{QIC}    & \textbf{QTR}    & \textbf{QQR}  & \textbf{Avg.}   \\
            \hline
            BERT-base            & 62.1            & 54.0            & 55.4          & 69.2          & 83.0          & 84.3            & 60.0            & \textbf{84.7} & 69.0            \\
            BERT-wwm-ext-base    & 61.7            & 54.0            & 55.4          & \textbf{70.1} & 83.9          & 84.5            & 60.9            & 84.4          & 69.4            \\
            RoBERTa-wwm-ext-base & 62.4            & 53.7            & 56.4          & 69.4          & 83.7          & 85.5            & 60.3            & 82.7          & 69.3            \\
            PCL-MedBERT          & 60.6            & 49.1            & 55.8          & 67.8          & 83.8          & 84.3            & 59.3            & 82.5          & 67.9            \\
            MacBERT-base         & 60.7            & 53.2            & \textbf{57.7} & 67.7          & \textbf{84.4} & 84.9            & 59.7            & 84.0          & 69.0            \\
            \modelname{} (ours)  & \textbf{62.984} & \textbf{54.932} & 56.484        & 69.655        & 83.632        & \textbf{85.858} & \textbf{61.099} & 84.587        & \textbf{69.904} \\
            \hline
            \multicolumn{10}{l}{Our \modelname{} achieves the highest average score among all competitors. The values are presented in percentage (\%).}
        \end{tabular}
    \end{center}
\end{table*}

\begin{table*}[t]
    \caption{Evaluation of pretraining techniques.\label{table:pretraining_techniques}}
    \renewcommand\arraystretch{1.2}
    \begin{center}
        \begin{tabular}{|l|c|c|c|c|c|c|c|c|c|c|}
            \hline
            \textbf{Techniques} & \textbf{Trainable Parameters} & \textbf{CMeEE}  & \textbf{CMeIE}  & \textbf{CDN}    & \textbf{CTC}    & \textbf{STS}    & \textbf{QIC}    & \textbf{QTR}    & \textbf{QQR}    & \textbf{Avg.}   \\
            \hline
            Fine-Tuning                     & 102M(100\%)  & 61.000          & 50.847          & 54.738           & 66.154          & 79.618          & 81.004          & 55.243          & 75.879          & 65.560             \\
            FL-Tuning$_{fl=1024}$           & 19M(15.59\%) & 62.147          & 52.977          & 56.136           & 68.308          & 83.033          & 84.805          & \textbf{62.892} & 83.522          & 69.275             \\
            LoRA$_{r=128}$                  & 21M(17.19\%) & 61.747          & 36.333          & 49.859           & 62.150          & 82.177          & 80.894          & 60.586          & 80.954          & 64.338             \\
            \methodname{} (ours)            & 21M(17.21\%) & \textbf{62.984} & \textbf{54.932} & \textbf{56.484}  & \textbf{69.655} & \textbf{83.632} & \textbf{85.858} & 61.099          & \textbf{84.587} & \textbf{69.904}    \\
            \hline
        \end{tabular}
    \end{center}
\end{table*}

We compare models by applying them to the downstream NLP tasks, specifically the Chinese Biomedical Language Understanding Evaluation (CBLUE~\cite{bib:cblue}) benchmark.
As shown in Table~\ref{table:finetuning_tasks}, the benchmark divides eight tasks into six categories: medical named entity recognition (NER), medical relationship extraction (RE), medical diagnosis normalization (NORM), medical textual similarity (TS), text classification (TC), and natural language inference (NLI).

\begin{itemize}
  \item \textbf{Named Entity Recognition (NER)} is the task of tagging entities in text with their corresponding type.
        CMeEE task provides a pre-defined schema including nine entity categories.
  \item \textbf{Relationship Extraction (RE)} aims to extract semantic relationships between two or more entities of a certain type from unstructured text into a number of semantic categories.
        There are 53 relations defined in the CMeIE task, including 10 synonymous sub-relationships and 43 other sub-relationships.
  \item \textbf{Lexical Normalization (NORM)} is the task of transforming a non-standard text into a standard register.
        Clinically, there might be up to hundreds of different synonyms for the same diagnosis, symptoms, or procedures.
        The task aims to find the standard phrases for the given clinical term.
  \item \textbf{Textual Similarity (TS)} deals with determining how semantically similar two pieces of text are.
        The CHIP-STS task aims to evaluate the generalization ability between disease types on Chinese disease questions and answer data.
  \item \textbf{Text Classification (TC)} includes simple short text classification and sentence intent classification.
        For CHIP-CTC, the task is to classify clinical trials' eligibility criteria.
        For KUAKE-QIC, the task is to classify each of them into one of 11 pre-defined medical intent categories.
  \item \textbf{Natural Language Inference (NLI)} is the task of determining whether a hypothesis is true (entailment) or false (contradiction) or undetermined (neutral) given a premise.
        For the KUAKE-QTR and KUAKE-QQR tasks, the dataset is used to estimate the relevance between the query and the title or another query.
\end{itemize}

For the named entity recognition task, we use BIO notation~\cite{bib:bio}, which differentiates the beginning (B), the inside (I) of entities, and the outside (O) of entities.
For the relationship extraction task, we split it into two steps: first, we recognize the subject and object entities and then extract the semantic relationships between them.
For the lexical normalization task, we use the recall and ranking approach.
For other tasks, we use simple classification architecture to predict the result.
The evaluation process is similar to the CBLUE benchmark toolkit~\cite{bib:cblue}, and we leverage the hyperparameters provided by the CBLUE RoBERTa baseline.

\subsection{Main Results}

In this section, we present the comparative analysis of different PLMs and pretraining techniques. We summarize the main findings and highlight the performance differences observed among the models and techniques.

\subsubsection{Comparison of PLMs}

For comparison, we use the CBLUE's public results of BERT-base, BERT-wwm-ext-base, RoBERTa-wwm-ext-base, PCL-MedBERT, and MacBERT-base baseline.
Table~\ref{table:benchmark_results} shows the performances of Chinese NLP tasks of the CBLUE benchmark.

For the NER task, our \modelname{} achieve a score of 62.984\%, outperforming the other models.
This demonstrates its ability to accurately extract biomedical entities from unstructured text.

In the RE task, we observe that \modelname{} outperforms the backbone model by about 1.2\%, surpassing the other models as well.
This highlights its effectiveness in extracting semantic relationships between entities of a certain type, demonstrating its superior performance in biomedical information extraction.

Regarding the NORM, TS, and TC tasks, \modelname{} achieves similar performance compared to the backbone model.
Notably, our model outperforms the other models in the QIC task, showcasing its advantage in accurately classifying medical intents.

In the NLI tasks, our model achieves the best performance, surpassing the backbone model by approximately 1.9\%. This signifies its ability to determine the logical relationships between query and title, highlighting its superiority in resolving the challenges for search engines.

Overall, our \modelname{} demonstrates strong performance across multiple task types and performs the best among models on average.
It outperforms the backbone model RoBERTa-wwm-ext-base and medically domain-specific model PCL-MedBERT by 0.6\%, and 2\%, respectively.
While achieving similar performance in NORM, TS, and TC tasks, our model stands out in the NER, RE, and NLI tasks.
These results highlight the versatility and effectiveness of our \modelname{} in addressing a wide range of biomedical language understanding challenges.

\subsubsection{Comparison of Pretraining Techniques}\label{experiment:ablation}

To investigate the impact of pretraining techniques on domain-specific models, we conduct several experiments using Chinese RoBERTa-wwm-ext-base as the backbone model.
With the original vocabulary and the same corpus as \modelname{}, we continually pretrain the model using Fine-Tuning and other methods.
To ensure a fair comparison, we adjust the hyperparameters of the techniques, excluding Fine-Tuning, to achieve a similar size of trainable parameters.
The evaluation results of the pretraining techniques are presented in Table~\ref{table:pretraining_techniques}.

The experimental results highlight the superior performance of the pretrained model using \methodname{}.
Compared to the Fine-Tuning approach, \methodname{} achieves remarkable results while training only about 17\% of the model parameters.
By adding additional parameters, \methodname{} consistently outperforms LoRA by an average margin of 5.5\%.
Notably, our experiments reveal that the Fine-Tuning method leads to suboptimal performance.
These findings suggest that continuing pretraining without considering domain-specific knowledge and representations may fail to effectively balance domain-specific information with general-domain representations.

\begin{figure}[!t]
  \small
  \centering
  \includegraphics[width=0.85\columnwidth]{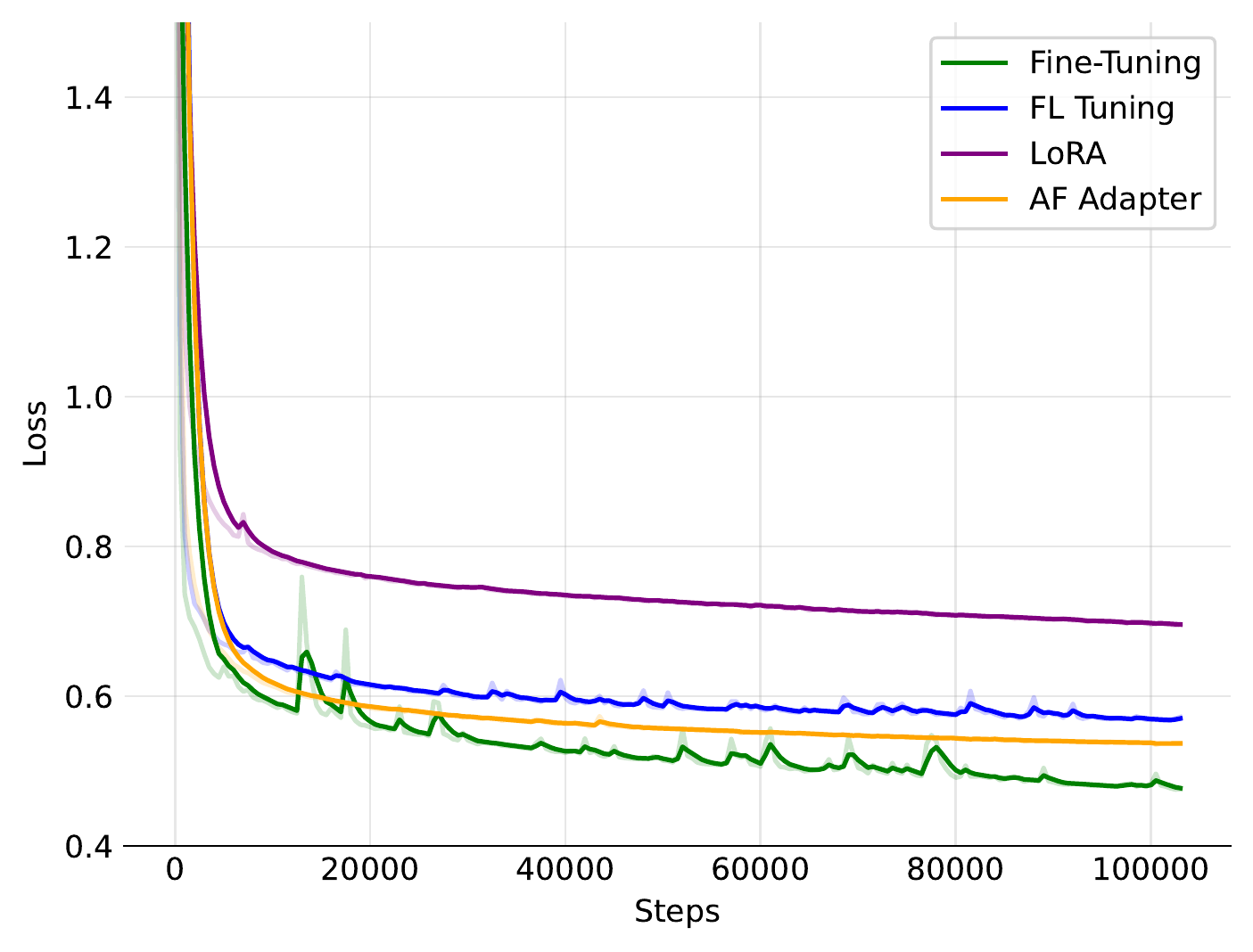}
  \caption{Convergence comparison of Fine-Tuning, FL-Tuning, LoRA and \methodname{}.
    The curves in dark color are obtained by smoothing the loss curves (light color).
    The smoothing function is $\alpha * previous\_smoothed\_value +(1-\alpha)*current\_value$, where $\alpha=0.6$ is the smooth weight.}
  \label{fig:loss}
\end{figure}

\subsection{Detailed Analysis}

In this subsection, we delve into a detailed analysis of the convergence and stability of the pretraining techniques in our study.
Additionally, we conduct an analysis of catastrophic forgetting, which assesses the extent each method retains previously learned knowledge while adapting to a new domain.


\textbf{Convergence and Stability Analysis.} We examine the convergence and stability of pretraining techniques throughout the pretraining process and present the results in Fig.~\ref{fig:loss}.

From the results, we observe that our \methodname{} exhibits a significant decrease in loss compared to LoRA.
This slower convergence indicates that the model requires more iterations to reach an optimal solution.
One possible explanation for this observation is that \methodname{} incorporates a ``parallel'' structure and the additional parameters enable the model to better learn and focus on the domain-specific information.

Alongside convergence, our \methodname{} demonstrates consistent and stable performance, maintaining a smooth and gradual decrease in loss.
In contrast, FL-Tuning and Fine-Tuning exhibit more fluctuations and variations in loss values.
This stability analysis further supports the robustness of our \methodname{} approach in learning domain-specific information.


\textbf{Catastrophic Forgetting Analysis.} To assess the forgetting problem caused by sequential task training, we randomly choose 10k general-domain samples from the ``WuDao'' corpus~\cite{bib:wudao}.
These samples are preprocessed using a masking rate of 15\%, ensuring consistent inputs across all compared models.
We evaluate the accuracy between Chinese RoBERTa-wwm-ext-base, Fine-Tuning based, FL-Tuning based, LoRA based, and \modelname{} trained in Section~\ref{experiment:ablation}.
The experimental result, presented in Table~\ref{table:catastrophic_forgetting}, demonstrates the models' performance in terms of catastrophic forgetting.
Our model achieves only -5.324\% accuracy reduction compared to the original RoBERTa.
Additionally, it is evident that \modelname{} outperforms Fine-Tuning based RoBERTa by a significant margin, achieving an accuracy of 82.428 compared to 70.964.
This indicates that our model better retains previously learned knowledge when trained on new tasks.
The model trained using fine-tuning exhibits a substantial drop in accuracy, indicating a significant forgetting of general knowledge.
This aligns with its overall poor performance in CBLUE, as shown in Table~\ref{table:pretraining_techniques}.

\begin{table}[tb]
    \caption{Catastrophic forgetting analysis.\label{table:catastrophic_forgetting}}
    \renewcommand\arraystretch{1.2}
    \begin{center}
        \begin{tabular}{|c|c|c|}
            \hline
            \textbf{Model}      & \textbf{Accuracy} & \textbf{Diff} \\
            \hline
            RoBERTa                    & 87.752            & -0            \\
            Fine-Tuning based RoBERTa  & 70.964            & -16.788       \\
            FL-Tuning based RoBERTa    & 83.555            & -4.197        \\
            LoRA based RoBERTa         & 78.231            & -9.521        \\
            \modelname{} (ours)        & 82.428            & -5.324        \\
            \hline
        \end{tabular}
    \end{center}
\end{table}

\section{Conclusion}

In this paper, we propose a continual pretraining approach, named \methodname{}, for building a domain-specific pretrained language model.
By incorporating additional attention heads and hidden units within the BERT-based model, \methodname{} enables effective learning of domain-specific knowledge while leveraging the general language representations.
We also present a Chinese biomedical-domain pretrained language model called \modelname{}, which is trained using \methodname{} and biomedical corpus.
The evaluation of \modelname{} on the CBLUE benchmark showcases its superior performance, outperforming other models of ``similar size'' by an average of 0.98\%.
Moreover, our analysis reveals that \methodname{} exhibits robust convergence and stability compared to other pretraining techniques.
These findings highlight the effectiveness and potential of \methodname{} in domain-specific pretraining for language models.

\section*{Acknowledgment}

The authors thank the anonymous reviewers for their valuable suggestions.
This work is supported by the National Key Research and Development Program of China (2021YFC2701800, 2021YFC2701801).

\bibliographystyle{unsrt}
\bibliography{reference}

\end{document}